\documentclass[11pt]{article}
\usepackage{url}
\makeatletter
\newcommand{\@BIBLABEL}{\@emptybiblabel}
\newcommand{\@emptybiblabel}[1]{}
\makeatother
\usepackage[hidelinks]{hyperref}
\usepackage[space]{cite}

\usepackage{acl2012}
\usepackage{times}
\usepackage{latexsym}
\usepackage[fleqn]{amsmath}
\usepackage{amssymb}
\usepackage{multirow}

\usepackage[utf8]{inputenc}
\usepackage{graphicx}
\usepackage{qtree}
\usepackage{bbm}
\usepackage{tikz}
\usepackage{verbatim}
\usepackage{amsmath}
\usepackage[noend]{algpseudocode}
\usepackage{algorithm}
\usepackage[T1]{fontenc}
\usepackage{tikz-dependency}
\usepackage{adjustbox}
\usepackage{tabularx}
\usepackage{enumitem}
\setitemize{noitemsep,topsep=0pt,parsep=0pt,partopsep=0pt}

\makeatletter
\renewcommand{\Function}[2]{%
  \csname ALG@cmd@\ALG@L @Function\endcsname{#1}{#2}%
  \def\jayden@currentfunction{#1}%
}
\newcommand{\funclabel}[1]{%
  \@bsphack
  \protected@write\@auxout{}{%
    \string\newlabel{#1}{{\jayden@currentfunction}{\thepage}}%
  }%
  \@esphack
}

\makeatother

\algrenewcommand\algorithmicindent{1.0em}%

\newcommand\R{\ensuremath{\mathbb{R}}}

\DeclareMathOperator*{\argmax}{arg\,max}
\title{Simple and Accurate Dependency Parsing \\ Using Bidirectional LSTM
Feature Representations}

\author{Eliyahu Kiperwasser \\
  Computer Science Department \\
  Bar-Ilan University \\
  Ramat-Gan, Israel \\
  {\tt elikip@gmail.com} \\\And
  Yoav Goldberg \\
  Computer Science Department \\
  Bar-Ilan University \\
  Ramat-Gan, Israel \\
  {\tt yoav.goldberg@gmail.com} \\}

\date{}

\begin{document}
\maketitle
\begin{abstract}

    We present a simple and effective scheme for dependency parsing which is
    based on bidirectional-LSTMs (BiLSTMs).
    Each sentence token is associated with a \mbox{BiLSTM} vector representing the
    token in its sentential context, and feature vectors are constructed by
    concatenating a few \mbox{BiLSTM} vectors.  The \mbox{BiLSTM} is trained jointly with the parser
    objective, resulting in very effective feature extractors for parsing.
    We demonstrate the effectiveness of the approach by applying it to a greedy
    transition-based parser as well as to a globally optimized graph-based
    parser.
    The resulting parsers have very simple architectures, and match or surpass the
    state-of-the-art accuracies on English and Chinese.

\end{abstract}

\section{Introduction}

The focus of this paper is on feature representation for dependency parsing,
using recent techniques from the neural-networks (``deep learning'') literature.
Modern approaches to dependency parsing can be broadly categorized into
graph-based and transition-based parsers \cite{kubler2008dependency}.
Graph-based parsers \cite{mcdonald-phd} treat parsing as a search-based structured prediction
problem in which the goal is learning a scoring function over dependency trees
such that the correct tree is scored above all other trees.  Transition-based
parsers \cite{arc-eager,nivre2008algorithms} treat parsing as a sequence of actions that produce a parse tree,
and a classifier is trained to score the possible actions at each stage of the
process and guide the parsing process.
Perhaps the simplest graph-based parsers are arc-factored (first
order) models \cite{mcdonald-phd}, in which the scoring function for a tree
decomposes over the individual arcs of the tree.  More elaborate models look at
larger (overlapping) parts, requiring more sophisticated inference and training
algorithms \cite{martins2009ilp,koo2010thirdorder}.  The basic transition-based parsers work in a greedy
manner, performing a series of locally-optimal decisions, and boast very fast
parsing speeds.  More advanced transition-based parsers introduce some search into the process
using a beam \cite{tale-two-parsers} or dynamic programming \cite{huang-sagae:2010:ACL}.

Regardless of the details of the parsing framework being used, a crucial step in
parser design is choosing the right \emph{feature function} for the underlying
statistical model.
Recent work (see Section \ref{sec:related} for an overview)
attempt to alleviate parts of the feature function design problem by moving from linear
to non-linear models, enabling the modeler to focus on a small set of ``core''
features and leaving it up to the machine-learning machinery to come up with good feature
combinations
\cite{chen2014fast,pei2015effective,lei-EtAl:2014:P14-1,taubtabib2015template}.
However, the need to carefully define a set of core features remains. For
example, the work of \newcite{chen2014fast} uses 18 different elements in its
feature function, while the work of \newcite{pei2015effective} uses 21 different
elements. Other works, notably \newcite{dyer2015transitionbased} and \newcite{le2014insideoutside}, propose more
sophisticated feature representations, in which the feature engineering is
replaced with architecture engineering.

In this work, we suggest an approach which is much simpler in terms of both
feature engineering and architecture engineering.
Our proposal (Section \ref{sec:method}) is centered around BiRNNs
\cite{irsoy2014opinion,schuster1997bidirectional}, and more specifically BiLSTMs
\cite{graves2008supervised},
which are strong and trainable sequence models (see Section \ref{subsec:birnn}).
The BiLSTM excels at representing elements in a sequence (i.e., words) together with
their contexts, capturing the element and an ``infinite'' window around
it.  We represent each word by its \mbox{BiLSTM} encoding, and use a concatenation of
a minimal set of such BiLSTM encodings as our feature function, which is then passed to a
non-linear scoring function (multi-layer perceptron). Crucially, the BiLSTM is
trained with the rest of the parser in order to learn a good feature
representation for the parsing problem.
If we set aside the inherent complexity of the \mbox{BiLSTM} itself and treat it as a
black box, our proposal results in a pleasingly simple feature extractor.

We demonstrate the effectiveness of the approach by using the BiLSTM feature
extractor in two parsing architectures, transition-based (Section \ref{sec:archybrid}) as well as a
graph-based (Section \ref{sec:mst}).
In the graph-based parser, we jointly train a structured-prediction model on top of a
BiLSTM, propagating errors from the structured objective all the way back to
the \mbox{BiLSTM} feature-encoder.  To the best of our knowledge, we are the first to
perform such end-to-end training of a structured prediction model and a
recurrent feature extractor for non-sequential outputs.\footnote{Structured
training of sequence tagging models over RNN-based representations was explored by
\newcite{chiu2015named} and \newcite{lample2016neural}.}

Aside from the novelty of the BiLSTM feature extractor and the end-to-end
structured training, we rely on existing models and techniques from
the parsing and structured prediction literature.
We stick to the simplest
parsers in each category -- greedy inference for the transition-based
architecture, and a first-order, arc-factored model for the graph-based
architecture.
Despite the simplicity of the parsing architectures and the
feature functions, we achieve near state-of-the-art parsing accuracies in both English
(93.1 UAS) and
Chinese (86.6 UAS), using a first-order parser with two features and while
training solely on Treebank data, without relying on semi-supervised signals such as pre-trained word
embeddings \cite{chen2014fast}, word-clusters \cite{koo2008semisup}, or
techniques such as tri-training \cite{weiss2015structured}.  When also including
pre-trained word embeddings, we obtain further improvements, with accuracies of
93.9 UAS (English) and 87.6 UAS (Chinese) for a greedy transition-based parser
with 11 features, and 93.6 UAS (En) / 87.4 (Ch) for a greedy transition-based
parser with 4 features.

\section{Background and Notation}
\paragraph{Notation} We use $x_{1:n}$ to denote a sequence of $n$ vectors $x_1,\cdots,x_n$.
$F_{\theta}(\cdot)$ is a function parameterized with parameters $\theta$.
We write $F_{L}(\cdot)$ as shorthand for $F_{\theta_L}$ -- an instantiation of $F$
with a specific set of parameters $\theta_L$.  We use $\circ$ to denote a vector
concatenation operation, and $v[i]$ to denote an indexing operation taking the
$i$th element of a vector $v$.

\subsection{Feature Functions in Dependency Parsing}
\label{subsec:feattemp}
Traditionally, state-of-the-art parsers rely on linear models over hand-crafted
feature functions.  The feature functions look at core components (e.g.
``word on top of stack'', ``leftmost child of the second-to-top word on the
stack'', ``distance between the head and the modifier words''), and are
comprised of several templates, where each template instantiates a binary
indicator function over a conjunction of core elements (resulting in features of
the form ``word on top of stack is X and leftmost child is Y and \ldots'').  The design of
the feature function -- which components to consider and which combinations of
components to include -- is a major challenge in parser design.
Once a good feature function is proposed in a paper it is usually adopted in later works,
and sometimes tweaked to improve performance.  Examples of good feature
functions are the feature-set proposed by \newcite{zhang11acl} for
transition-based parsing (including roughly 20 core components and 72 feature
templates), and the feature-set proposed by \newcite{mst} for graph-based parsing, with the paper listing 18
templates for a first-order parser, while the first order feature-extractor in the actual implementation's code (MSTParser\footnote{\url{http://www.seas.upenn.edu/~strctlrn/MSTParser/MSTParser.html}}) includes roughly a hundred feature templates.

The core features in a transition-based parser usually look at information such as the word-identity and
part-of-speech (POS) tags of a fixed number of words on top of the stack, a fixed
number of words on the top of the buffer, the modifiers (usually left-most and
right-most) of items on the stack and on the buffer, the number of modifiers of
these elements, parents of words on the stack, and the length of the spans spanned by the words on the stack.
The core features of a first-order graph-based parser usually take into account
the word and POS of the head and modifier items, as well as POS-tags of
the items around the
head and modifier, POS tags of items between the head and modifier,
and the distance and direction between the head and modifier.

\subsection{Related Research Efforts}
\label{sec:related}
Coming up with a good feature-set for a parser is a hard and time consuming
task, and many researchers attempt to reduce the required manual effort.
The work of \newcite{lei-EtAl:2014:P14-1} suggests a low-rank tensor
representation to automatically find good feature combinations.
\newcite{taubtabib2015template} suggest a kernel-based approach to
implicitly consider all possible feature combinations over sets of
core-features.  The recent popularity of neural networks prompted a move from
templates of sparse, binary indicator features to dense core feature encodings fed into non-linear
classifiers. \newcite{chen2014fast} encode each core feature
of a greedy transition-based parser as a dense low-dimensional vector, and the vectors are then concatenated
and fed into a non-linear classifier (multi-layer perceptron) which can
potentially capture arbitrary feature combinations.  \newcite{weiss2015structured} showed further gains using the same approach
coupled with a somewhat improved set of core features, a more involved network
architecture with skip-layers, beam search-decoding, and careful hyper-parameter
tuning.  \newcite{pei2015effective} apply a similar methodology to
graph-based parsing.
While the move to neural-network classifiers alleviates the need for
hand-crafting feature-combinations, the need to carefully define a set of core
features remain. For example, the feature representation  in \newcite{chen2014fast}
is a concatenation of 18 word vectors, 18 POS vectors and 12 dependency-label
vectors.%, while the feature representation in \cite{pei2015effective} is a concatenation of @@\ldots.
\footnote{In all of these neural-network based approaches, the vector representations of
words were initialized using pre-trained word-embeddings derived from a large
corpus external to the training data.  This puts the approaches in the
semi-supervised category, making it hard to tease apart the contribution of the
automatic feature-combination component from that of the semi-supervised
component.}

The above works tackle the effort in hand-crafting effective
feature combinations.
A different line of work attacks the feature-engineering problem by suggesting novel
neural-network architectures for encoding the parser state, including
intermediately-built subtrees, as vectors which are then fed to non-linear
classifiers.  
Titov and Henderson encode the parser state using incremental sigmoid-belief
networks \shortcite{titov-henderson:2007:IWPT2007}.
In the work of \newcite{dyer2015transitionbased}, the entire stack and buffer of a
transition-based parser are encoded as a stack-LSTMs, where each stack element
is itself based on a compositional representation of parse trees.  \newcite{le2014insideoutside} encode each tree node as two compositional representations capturing the
inside and outside structures around the node, and feed the representations into
a reranker.  A similar reranking approach, this time based on convolutional
neural networks, is taken by
\newcite{zhu2015reranking}.
Finally, in \newcite{kiperwasser2016ef} we 
present an Easy-First parser based on a novel hierarchical-LSTM tree encoding.

In contrast to these, the approach we present in this work results in much
simpler feature functions, without resorting to elaborate network architectures or compositional tree representations.

Work by \newcite{vinlays2014grammar} employs a
sequence-to-sequence with attention architecture for constituency parsing.
Each token in the input sentence is encoded in a deep-BiLSTM representation, and
then the tokens are fed as input to a deep-LSTM that predicts a sequence of
bracketing actions based on the already predicted bracketing as well as the encoded BiLSTM
vectors. A trainable attention mechanism is used to guide the parser to relevant
\mbox{BiLSTM} vectors at each stage.  This architecture shares with ours the use of
BiLSTM encoding and end-to-end training.  The sequence of bracketing actions
can be interpreted as a sequence of Shift and Reduce operations of a transition-based parser. However, while the
parser of Vinyals et al.\ relies on a trainable attention mechanism for focusing on specific 
BiLSTM vectors, parsers in the transition-based family we use in Section
\ref{sec:archybrid} use a human designed stack and buffer mechanism to manually direct the
parser's attention.  While the effectiveness of the trainable attention approach
is impressive, the stack-and-buffer guidance of transition-based parsers results
in more robust learning.
Indeed, work by \newcite{cross2016incremental}, published while working on the camera-ready
version of this paper, show that the same methodology as ours is highly effective also for
greedy, transition-based constituency parsing, surpassing the beam-based architecture of
Vinyals et al. (88.3F vs. 89.8F points) when trained on the Penn Treebank dataset and without
using orthogonal methods such as ensembling and up-training.

\subsection{Bidirectional Recurrent Neural Networks}
\label{subsec:birnn}
Recurrent neural networks (RNNs) are statistical
learners for modeling sequential data. An RNN allows one to model the $i$th element in the sequence
based on the past -- the elements $x_{1:i}$ up to and including it. The RNN model provides a framework for
conditioning on the entire history $x_{1:i}$ without resorting to the Markov assumption
which is traditionally used for modeling sequences. RNNs were shown to be
capable of learning to count, as well as to model line lengths and complex
phenomena such as bracketing and code indentation \cite{karpathy2015visualizing}.
Our proposed feature extractors are based on a bidirectional recurrent neural network
(BiRNN), an extension of RNNs that take into account both the past $x_{1:i}$ and the
future $x_{i:n}$.  We use a specific flavor of RNN called a
long short-term memory network (LSTM).
For brevity, we treat RNN as an abstraction, without getting
into the mathematical details of the implementation of the RNNs and LSTMs.  For
further details on RNNs and LSTMs, the reader is referred to
\newcite{goldberg-primer} and \newcite{cho-primer}.

The recurrent neural network (RNN) abstraction is a parameterized function
$\textsc{Rnn}_\theta(x_{1:n})$
mapping a sequence of $n$ input vectors $x_{1:n}$, $x_i \in \R^{d_{in}}$ to a
sequence of $n$ output vectors $h_{1:n}, h_i \in \R^{d_{out}}$.  Each output vector $h_i$ is conditioned
on all the input vectors $x_{1:i}$, and can be thought of as a \emph{summary} of
the prefix $x_{1:i}$ of $x_{1:n}$.  In our notation, we ignore the intermediate
vectors $h_{1:{n-1}}$ and take the output of $\textsc{Rnn}_\theta(x_{1:n})$ to
be the vector $h_n$.

A \emph{bidirectional RNN} is composed of two RNNs, $\textsc{Rnn}_{F}$ and
$\textsc{Rnn}_{R}$, one
reading the sequence in its regular order, and the other reading it in reverse.
Concretely, given a sequence of vectors $x_{1:n}$ and a desired index $i$, the function
$\textsc{BiRnn}_{\theta}(x_{1:n}, i)$ is defined as:
\begin{align*}
    \textsc{BiRnn}_{\theta}(x_{1:n},i) &= \textsc{Rnn}_{F}(x_{1:i}) \circ
    \textsc{Rnn}_{R}(x_{n:i})
\end{align*}
The vector $v_i = \textsc{BiRnn}(x_{1:n}, i)$ is then a representation of the $i$th item
in $x_{1:n}$, taking into account both the entire history $x_{1:i}$ and the
entire future $x_{i:n}$ by concatenating the matching $\textsc{Rnn}$s.  We can view the BiRNN encoding of an item $i$ as
representing the item $i$ together with a context of an infinite window around
it.

\paragraph{Computational Complexity}
Computing the BiRNN vectors encoding of the $i$th element of a sequence
$x_{1:n}$ requires $O(n)$ time for computing the two RNNs and concatenating
their outputs.  A naive approach of computing the bidirectional
representation of all $n$ elements result in $O(n^2)$ computation.
However, it is trivial to compute the BiRNN encoding of all sequence items in
linear time by pre-computing $\textsc{RNN}_F(x_{1:n})$ and
$\textsc{RNN}_R(x_{n:1})$, keeping the intermediate representations, and
concatenating the required elements as needed.

\paragraph{BiRNN Training}
Initially, the BiRNN encodings $v_i$ do not capture any particular information.
During training, the encoded vectors $v_i$ are fed into further network layers,
until at some point a prediction is made, and a loss is incurred. The
back-propagation algorithm is used to compute the gradients of all the
parameters in the network (including the BiRNN parameters) with respect to the
loss, and an optimizer is used to update the parameters according to the
gradients. The training procedure causes the BiRNN function to extract from the
input sequence $x_{1:n}$ the relevant information for the task task at hand.

\paragraph{Going deeper}
We use a variant of \emph{deep bidirectional RNN} (or $k$-layer BiRNN) which is composed of $k$ BiRNN functions
$\textsc{BiRnn}_{1},\cdots,\textsc{BiRnn}_{k}$ that feed into each other: the output
$\textsc{BiRnn}_\ell(x_{1:n},1),\dots,\textsc{BiRnn}_\ell(x_{1:n}, n)$ of $\textsc{BiRnn}_{\ell}$ becomes the input of $\textsc{BiRnn}_{\ell+1}$.  Stacking BiRNNs in this way has been empirically shown to be effective \cite{irsoy2014opinion}.  
In this work, we use
BiRNNs and deep-BiRNNs interchangeably, specifying the number of layers when needed.

\paragraph{Historical Notes}
RNNs were introduced by \newcite{elman1990finding}, and extended to BiRNNs by
\newcite{schuster1997bidirectional}.  The LSTM variant of RNNs is due to
\newcite{hochreiter1997long}.  \mbox{BiLSTMs} were recently popularized by \newcite{graves2008supervised}, and deep
BiRNNs were introduced to NLP by \newcite{irsoy2014opinion}, who
used them for sequence tagging. 
In the context of parsing, \newcite{lewis2016lstm} and \newcite{Vaswani:2016:NAACL} use a \mbox{BiLSTM} sequence tagging model to assign a CCG supertag for each token in the sentence.  \newcite{lewis2016lstm} feeds the resulting supertags sequence into an A* CCG parser. 
\newcite{Vaswani:2016:NAACL} adds an additional layer of LSTM which receives the \mbox{BiLSTM} representation together with the k-best supertags for each word and outputs the most likely supertag given previous tags, and then feeds the predicted supertags to a discriminitively trained parser. In both works, the \mbox{BiLSTM} is trained to produce accurate CCG supertags, and is not aware of the global parsing objective.
\section{Our Approach}
\label{sec:method}

We propose to replace the hand-crafted feature functions in favor of
minimally-defined feature functions which make use of automatically learned Bidirectional LSTM
representations.

Given $n$-words input sentence $s$ with words $w_1,\dots,w_n$ together with the
corresponding POS tags $t_1,\dots,t_n$,\footnote{
In this work the tag sequence is assumed to be given, and in practice is predicted by an external model. Future work will address relaxing
this assumption.} we associate each word $w_i$ and POS $t_i$ with embedding
vectors $e(w_i)$ and $e(t_i)$, and create a sequence of input vectors
$x_{1:n}$ in which each
$x_i$ is a concatenation of the corresponding word and POS vectors:
\begin{align*}
    x_i = e(w_i) \circ e(p_i)
\end{align*}
The embeddings are trained together with the model.
This encodes each word in isolation, disregarding its context. We introduce
context by representing each input element as its (deep) \mbox{BiLSTM} vector, $v_i$:
\begin{align*}
    v_i &= \textsc{BiLstm}(x_{1:n},i)
\end{align*}
Our feature function $\phi$ is then a concatenation of a small number of
BiLSTM vectors.  The exact feature function is parser dependent and will be
discussed when discussing the corresponding parsers.  The resulting feature
vectors are then scored using a non-linear function, namely a multi-layer
perceptron with one hidden layer (MLP):
\[MLP_\theta(x) = W^2\cdot \tanh(W^1\cdot x + b^1) + b^2\]
where
$\theta=\{W^1,W^2,b^1,b^2\}$ are the model parameters.

Beside using the BiLSTM-based feature functions, we make use of standard parsing
techniques.  Crucially, the \mbox{BiLSTM} is trained jointly with the rest of the
parsing objective.  This allows it to learn representations which are suitable
for the parsing task.

Consider a concatenation of two \mbox{BiLSTM} vectors ($v_i \circ v_j$) scored using an MLP.
The scoring function has access to the words and POS-tags of $v_i$ and $v_j$, as
well as the words and POS-tags of the words in an infinite window surrounding
them.  As LSTMs are known to capture length and sequence position information, it is very
plausible that the scoring function can be sensitive also to the distance
between $i$ and $j$, their ordering, and the sequential material between them.

\paragraph{Parsing-time Complexity}
Once the \mbox{BiLSTM} is trained, parsing is performed by first computing the BiLSTM
encoding $v_i$ for each word in the sentence (a linear time operation).\footnote{
While the \mbox{BiLSTM} computation is quite efficient as it is, as demonstrated by \newcite{lewis2016lstm},
if using a GPU implementation the \mbox{BiLSTM} encoding
can be efficiently performed over many of sentences in parallel, making its computation cost
almost negligible.}
Then,
parsing proceeds as usual, where the feature extraction involves a concatenation
of a small number of the pre-computed $v_i$ vectors.

\section{Transition-based Parser}
\label{sec:archybrid}

\begin{figure*}[ht!]
\includegraphics[scale=0.55]{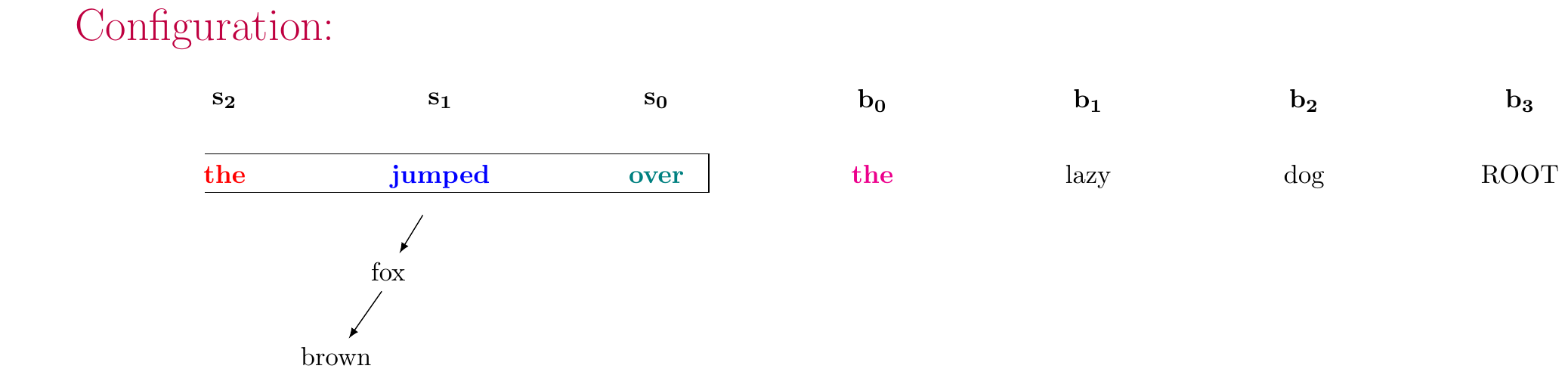}
\includegraphics[scale=0.55]{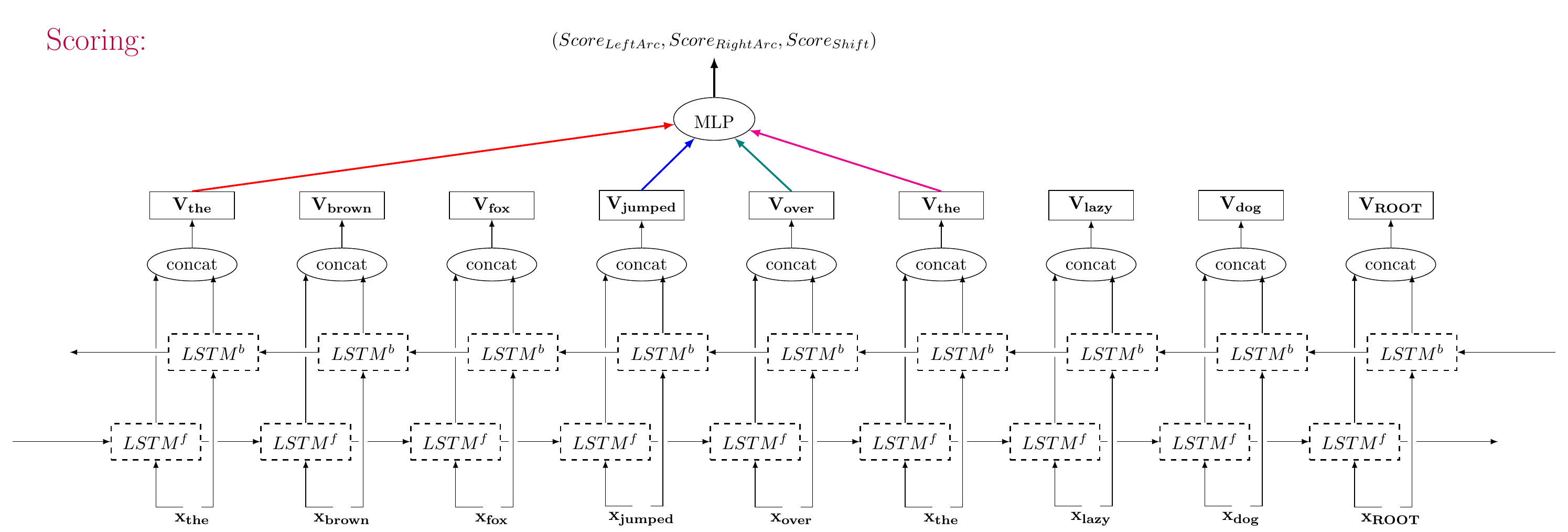}

% \vspace*{-2em}
\caption{\small Illustration of the neural model scheme of the transition-based parser when calculating the scores of the possible transitions in a given configuration.
The configuration (stack and buffer) is depicted on the top. Each transition is scored using an MLP that is fed the \mbox{BiLSTM} encodings of the first word in the buffer and the three words at the top of the stack (the colors of the words correspond to colors of the MLP inputs above), and a transition is picked greedily. Each $x_i$ is a concatenation of a word and a POS vector, and possibly an additional external embedding vector for the word. The figure depicts a single-layer BiLSTM, while in practice we use two layers. When parsing a sentence, we iteratively compute scores for all possible transitions and apply the best scoring action until the final configuration is reached.}
\label{fig:firstorder}
\end{figure*}

We begin by integrating the feature extractor in a transition-based parser
\cite{nivre2008algorithms}. We follow the notation in
\newcite{tacl2013dynamic}.
The transition-based parsing framework assumes a transition system, an abstract
machine that processes sentences and produces parse trees. The transition system
has a set of configurations and a set of transitions which are applied to
configurations. When parsing a sentence, the system is initialized to an initial
configuration based on the input sentence, and transitions are repeatedly
applied to this configuration. After a finite number of transitions, the system
arrives at a terminal configuration, and a parse tree is read off the terminal
configuration. In a greedy parser, a classifier is used to choose the transition
to take in each configuration, based on features extracted from the
configuration itself. The parsing algorithm is presented in Algorithm
\ref{alg:transition-parsing} below.

\begin{algorithm}[h]
   \caption{Greedy transition-based parsing}
   \label{alg:transition-parsing}
\begin{algorithmic}[1]
   \State \textbf{Input:} sentence $s = w_1,\dots,x_w,\;t_1,\dots,t_n$, parameterized function
   $\textsc{Score}_\mathbf{\theta}(\cdot)$ with parameters $\mathbf{\theta}$.
   \State $c \gets \textsc{Initial}(s)$\label{line:init}
   \While {\textbf{not} $\textsc{Terminal}(c)$}
   \State $\hat{t} \gets \argmax_{t\in\textsc{Legal}(c)}
   \textsc{Score}_\mathbf{\theta}\big(\phi(c), t\big)$\label{line:score}
   \State $c \gets \hat{t}(c)$  
   \EndWhile
   \State \Return $tree(c)$\label{line:final}
\end{algorithmic}
\end{algorithm}

Given a sentence $s$, the parser is initialized with the configuration $c$ (line
\ref{line:init}).
Then, a feature function $\phi(c)$ represents the configuration $c$ as
a vector, which is fed to a scoring function $\textsc{Score}$ assigning
scores to (configuration,transition) pairs. $\textsc{Score}$ scores the
possible transitions $t$, and the highest scoring transition $\hat{t}$ is chosen (line
\ref{line:score}).  The transition $\hat{t}$ is applied to the configuration,
resulting in a new parser configuration.  The process ends when reaching a final
configuration, from which the resulting parse tree is read and returned (line
\ref{line:final}).

Transition systems differ by the way they define configurations, and by the
particular set of transitions available to them.
A parser is determined by the choice of a transition system, a feature function $\phi$ and a scoring
function $\textsc{Score}$.  Our choices are detailed below.

\paragraph{The Arc-Hybrid System}
Many transition systems exist in the literature. In this work, we use the
arc-hybrid transition system \cite{kuhlmann2011}, which is similar to the more
popular arc-standard system \cite{arc-eager}, but for which an efficient
dynamic oracle is available \cite{coling2012dynamic,tacl2013dynamic}.
In the arc-hybrid system, a configuration $c = (\sigma, \beta, T)$ consists of a
stack $\sigma$, a buffer $\beta$, and a set $T$ of dependency arcs. 
Both the stack and the buffer hold integer indices pointing to sentence elements.
Given a sentence $s = w_1, \dots , w_n,\;t_1,\dots,t_n$, the system
is initialized with an empty stack, an empty arc set, and $\beta = 1, \dots ,
n,\mathtt{ROOT}$ , where $\mathtt{ROOT}$ is
the special root index. Any configuration $c$ with an empty stack and a buffer
containing only $\mathtt{ROOT}$ is
terminal, and the parse tree is given by the arc set $T_c$ of $c$.
The arc-hybrid system allows 3 possible transitions,
$\textsc{Shift}$, $\textsc{Left}_{\ell}$ and $\textsc{Right}_{\ell}$, defined
as:
\vspace{-5pt}
\begin{center}
\begin{scalebox}{0.8}{
\begin{tabular}{ll}
    $\textsc{Shift}[(\sigma,\;\; b_0 | \beta,\;\; T)]$ &$= (\sigma | b_0,\;\; \beta,\;\; T)$ \\
    $\textsc{Left}_{\ell}[(\sigma | s_1 | s_0 ,\;\; b_0 | \beta,\;\; T)]$ &$= (\sigma | s_1,\;\; b_0 | \beta,\;\; T \cup \{(b_0, s_0, \ell)\})$ \\
    $\textsc{Right}_{\ell}[(\sigma | s_1 | s_0 ,\;\; \beta,\;\; T)]$ &$= (\sigma | s_1,\;\; \beta,\;\; T \cup \{(s_1, s_0, \ell)\})$ \\
\end{tabular}}\end{scalebox}
\end{center}
\vspace{5pt}

\noindent The \textsc{Shift} transition moves the first item of the buffer ($b_0$) to the stack.
The \textsc{Left}$_\ell$ transition removes the first item on top of the stack
($s_0$) and attaches it as a modifier to $b_0$ with label $\ell$, adding the arc $(b_0,s_0,\ell)$.
The \textsc{Right}$_\ell$ transition removes $s_0$ from the stack and attaches
it as a modifier to the next item on the stack ($s_1$), adding the arc
$(s_1,s_0,\ell)$.

\paragraph{Scoring Function}
Traditionally, the scoring function $\textsc{Score}_\mathbf{\theta}(x,t)$ is a
discriminative linear model of the form $\textsc{Score}_W(x,t) =
(W\cdot x)[t]$.
The linearity of $\textsc{Score}$ required the feature function $\phi(\cdot)$ to encode
non-linearities in the form of combination features.  We
follow Chen and Manning \shortcite{chen2014fast} and replace the linear scoring
model with an MLP.
\[
\textsc{Score}_\mathbf{\theta}(x,t) = MLP_\mathbf{\theta}(x)[t]
\]

\paragraph{Simple Feature Function}
The feature function $\phi(c)$ is typically complex (see Section \ref{subsec:feattemp}).
Our feature function is the concatenated \mbox{BiLSTM} vectors of the top 3 items on the
stack and the first item on the buffer.
I.e., for a configuration $c = (\dots | s_2 | s_1 | s_0,\;\; b_0 | \dots,\;\; T)$
the feature extractor is defined as:
\begin{align*}
    \phi(c) &= v_{s_2}\circ v_{s_1} \circ v_{s_0} \circ v_{b_0} \\
    v_i &= \textsc{BiLstm}(x_{1:n}, i) 
\end{align*}

This feature function is rather minimal: it takes into account the BiLSTM
representations of $s_1, s_0$ and
$b_0$, which are the items affected by the possible transitions being scored, as
well as one extra
stack context $s_2$.\footnote{An additional buffer context is not needed, as
$b_1$ is by definition adjacent to $b_0$, a fact that we expect the BiLSTM
encoding of $b_0$ to capture.  In contrast, $b_0$, $s_0$, $s_1$ and $s_2$ are
not necessarily adjacent to each other in the original sentence.}
Figure 1 depicts transition scoring with our architecture and this feature function. Note that, unlike previous work, this feature function \emph{does not} take into account $T$, the already built
structure.  The high parsing accuracies in the experimental sections suggest
that the \mbox{BiLSTM} encoding is capable of estimating a lot of the missing
information based on the provided stack and buffer elements and the sequential
content between them.

While not explored in this work, relying on only four word indices for scoring
an action results in
very compact state signatures, making our proposed feature representation 
very appealing for use in transition-based parsers that employ dynamic-programming search \cite{huang-sagae:2010:ACL,kuhlmann2011}.

\paragraph{Extended Feature Function}
One of the benefits of the greedy transition-based parsing framework is
precisely its ability to look at arbitrary features from the already built tree.
If we allow somewhat less minimal feature function, we could add the \mbox{BiLSTM}
vectors corresponding to the right-most and left-most modifiers of $s_0$, $s_1$
and $s_2$, as well as the left-most modifier of $b_0$,
reaching a total of 11 \mbox{BiLSTM} vectors. We
refer to this as the \emph{extended feature set}. As we'll see in Section
\ref{sec:results}, using the extended set does indeed improve parsing accuracies when using
pre-trained word embeddings, but has a minimal effect in the fully-supervised
case.\footnote{We did not experiment with other feature configurations. It is well possible that not all of the additional 7 child encodings are needed for the observed accuracy gains, and that a smaller feature set will yield similar or even better improvements.}

\subsection{Details of the Training Algorithm}
The training objective is to set the score of correct transitions above the
scores of incorrect transitions.
We use a margin-based objective, aiming to
maximize the margin between the highest scoring correct action and the highest
scoring incorrect action. The \emph{hinge loss} at each parsing configuration
$c$ is defined as:
    \vspace{-1em}
\begin{center}
\begin{align*}
%max\{ 0, 1 -& max_{t_o \in G}Score(t_o, c)  \\
%           +& max_{t_p \in A\setminus G}Score(t_p, c)  \} = \\
max\Big( 0, 1 -& \max_{t_o \in G}MLP\big(\phi(c)\big)[t_o]  \\
           +& \max_{t_p \in A\setminus G}MLP\big(\phi(c)\big)[t_p]  \Big)
\end{align*}

\end{center}

%\begin{figure*}
%\input{trans.tex}
%\end{figure*}

\noindent where $A$ is the set of possible transitions and $G$ is the set of
correct (gold) transitions at the current stage. 
At each stage of the training process the parser scores the possible transitions
$A$, incurs a loss, selects a transition to follow, and moves to the next
configuration based on it.
The local losses are summed throughout the parsing
process of a sentence, and the parameters are updated with respect to the sum of the losses at
sentence boundaries.\footnote{To increase gradient stability and training speed,
we simulate mini-batch updates by only updating the parameters when the sum of local losses contains
at least 50 non-zero elements.  Sums of fewer elements are carried across
sentences.  This assures us a sufficient number of gradient samples for every update thus minimizing the effect of
gradient instability.}
The gradients of the entire network (including the MLP and the BiLSTM) with respect to the sum of the losses are
calculated using the backpropagation algorithm.
As usual, we perform several training iterations over the training corpus,
shuffling the order of sentences in each iteration.

\paragraph{Error-Exploration and Dynamic Oracle Training}
We follow \newcite{tacl2013dynamic};\newcite{coling2012dynamic} in
using error exploration training with a dynamic-oracle, which we briefly
describe below.

At each stage in the training process, the parser assigns scores to all the
possible transitions $t \in A$. It then selects a transition, applies it, and moves to the
next step. 
Which transition should be followed?  A common approach follows the highest
scoring transition that can lead to the gold tree. However, when training in
this way the parser sees only configurations that result from following correct
actions, and as a result tends to suffer from error propagation at test time.
Instead, in error-exploration training the parser follows the highest scoring
action in $A$ during training even if this action is incorrect, exposing it to
configurations that result from erroneous decisions.
This strategy requires defining the set $G$ such that the correct actions to
take are well-defined also for states that cannot lead to the gold tree.
Such a set $G$ is called a \emph{dynamic oracle}. We perform error-exploration
training using the dynamic-oracle defined by \newcite{tacl2013dynamic}.

\paragraph{Aggressive Exploration} We found that even when using error-exploration, after one iteration the model
remembers the training set quite well, and does not make enough errors to make
error-exploration effective.  In order to expose the parser to more errors, we 
follow an aggressive-exploration scheme: we sometimes follow incorrect transitions also if
they score below correct transitions.  Specifically, when the score of the correct
transition is greater than that of the wrong transition but the difference is smaller
than a margin constant, we chose to follow the incorrect action with probability
$p_{agg}$ (we use $p_{agg}=0.1$ in our experiments).

\paragraph{Summary}
The greedy transition-based parser follows standard techniques from the
literature (margin-based objective, dynamic oracle training, error exploration,
MLP-based non-linear scoring function).
We depart from the literature by replacing the hand-crafted feature function
over carefully selected components of the configuration with a concatenation of
BiLSTM representations of a few prominent items on the stack and the
buffer, and training the \mbox{BiLSTM} encoder jointly with the rest of the network. 

\section{Graph-based Parser}
\begin{figure*}
\begin{center}
\includegraphics[scale=0.7]{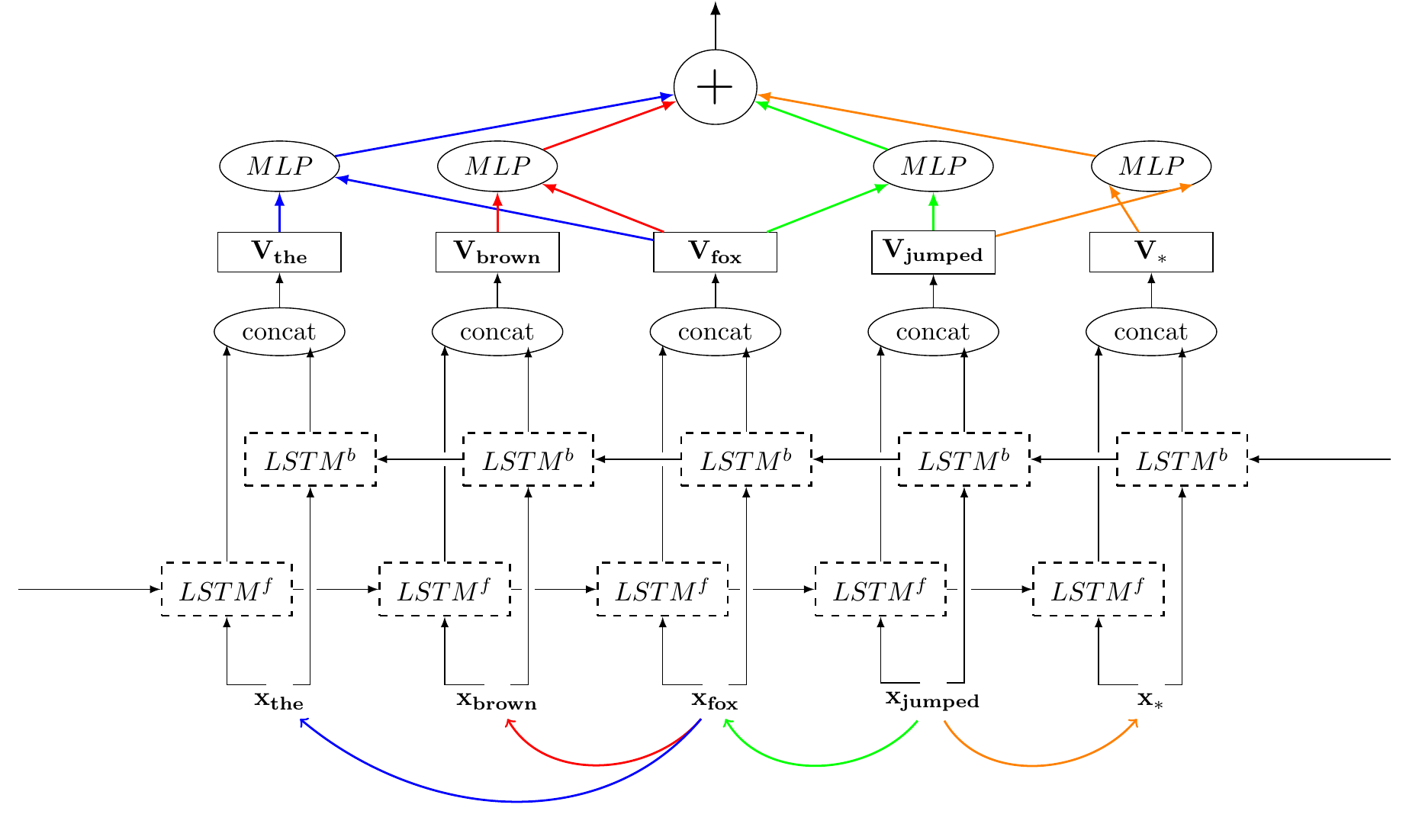}
\end{center}
\vspace*{-2em}
\caption{\small Illustration of the neural model scheme of the graph-based parser when calculating the score of a given parse tree.
The parse tree is depicted below the sentence. Each dependency arc in the sentence is scored using an MLP that is fed the \mbox{BiLSTM} encoding of the words at the arc's end points (the colors of the arcs correspond to colors of the MLP inputs above), and the individual arc scores are summed to produce the final score. All the MLPs share the same parameters. The figure depicts a single-layer BiLSTM, while in practice we use two layers. When parsing a sentence, we compute scores for all possible $n^2$ arcs, and find the best scoring tree using a dynamic-programming algorithm.}
\label{fig:firstorder}
\end{figure*}
\label{sec:mst}
Graph-based parsing follows the common structured prediction paradigm
\cite{taskar2005learning,mst}:
\begin{align*}
    predict(s) &= \argmax_{y \in \mathcal{Y}(s)} score_{global}(s,y) \\
    score_{global}(s,y) &= \sum_{part \in y} score_{local}(s, part) %\\
%score^{part}(x, part) &= w \cdot \phi(x, part)
\end{align*}

\noindent Given an input sentence $s$ (and the corresponding sequence of vectors
$x_{1:n}$) we look for the highest-scoring parse tree $y$ in the space
$\mathcal{Y}(s)$ of valid dependency trees over $s$. In order to make the search tractable, 
the scoring function is decomposed to the sum of local scores for each part independently. 

In this work, we focus on arc-factored graph based approach presented in \newcite{mst}.
Arc-factored parsing decomposes the score of a tree to the sum of the score of its
head-modifier arcs $(h,m)$:
\begin{align*}
parse(s) = \argmax_{y \in \mathcal{Y}(s)} \sum_{(h,m)\in y} score\big(\phi(s, h, m)\big)
\end{align*}
Given the scores of the arcs the highest scoring projective tree can be
efficiently found using Eisner's decoding algorithm \shortcite{eisner1996dep}.
McDonald et al.\ and most subsequent work estimate the local score of an arc by a
linear model parameterized by a weight vector $w$, and a feature function $\phi(s,h,m)$ assigning a sparse
feature vector for an arc linking modifier $m$ to head $h$. We follow \newcite{pei2015effective} and replace the linear scoring function with an
MLP.

The feature extractor $\phi(s,h,m)$ is usually complex, involving many elements (see  Section \ref{subsec:feattemp}).
In contrast, our feature extractor uses merely the \mbox{BiLSTM} encoding of the head word and the
modifier word:
\begin{align*}
%FV(h,m) = 
& \phi(s,h,m) = \textsc{BiRnn}(x_{1:n}, h) \circ \textsc{BiRnn}(x_{1:n}, m) %\\
%& BI-RNN(x, i) = RNN_F(w_1,\dots, w_i) \circ RNN_B(w_n,\dots,w_i)
%& BI-RNN(h) \circ BI-RNN(m) = \\
%& RNN_F(w_1,\dots, w_h) \circ RNN_B(w_n,\dots,w_h) \circ \\
%& RNN_F(w_1,\dots, w_m) \circ RNN_B(w_n,\dots,w_m)
\end{align*}

\noindent The final model is:
\begin{align*}
 parse(s) &= \argmax_{y \in \mathcal{Y}(s)} score_{global}(s,y) \\
          &= \argmax_{y \in \mathcal{Y}(s)} \sum_{(h, m) \in y} score\big(\phi(s,h,m)\big) \\
          &= \argmax_{y \in \mathcal{Y}(s)} \sum_{(h, m) \in y} MLP(v_h \circ v_m) \\
%& score_{arc}(x, h, m) = MLP(v_h \circ v_m) \\
%& score_{arc}(x, h, m) = MLP(v_h \circ v_m) \\
v_i &= \textsc{BiRnn}(x_{1:n}, i)
\end{align*}
\noindent The architecture is illustrated in Figure \ref{fig:firstorder}.

\paragraph{Training} The training objective is to set the score function such that correct tree $y$ is scored
above incorrect ones. We use a margin-based objective
\cite{mst,lecun2006tutorial}, 
aiming to maximize the margin between the score of the gold tree $y$ and the highest scoring incorrect
tree $y'$. We define a hinge loss with respect to a gold tree $y$ as:

\begin{align*}
    %max(0, 1 - \max_{y' \neq y} & \textsc{Score}_{global}(x,y') \\
    %+ & \textsc{Score}_{global}(x,y))  \\
    %= max(0, 1 - \max_{y' \neq y}&\sum_{(h,m) \in y'}{\textsc{Score}(\phi(x,h,m))} \\
    %+& \sum_{(h,m) \in y}\textsc{Score}(\phi(x,h,m))) \\
    max\Big(0, 1 - \max_{y' \neq y}&\sum_{(h,m) \in y'}{MLP(v_h \circ v_m)} \\
 +& \sum_{(h,m) \in y}{MLP(v_h \circ v_m)}\Big) 
 %\\
%& ) 
%v_i = \textsc{BiRnn}(x_{1:n}, i)
\end{align*}
\noindent Each of the tree scores is then calculated by activating the MLP on the arc
representations. The entire loss can viewed as the sum of multiple neural
networks, which is sub-differentiable. We calculate the gradients of the entire
network (including to the \mbox{BiLSTM} encoder and word embeddings).

\paragraph{Labeled Parsing}
Up to now, we described unlabeled parsing. 
A possible approach for adding labels is to score the combination of an
unlabeled arc $(h,m)$ and its label $\ell$ by considering the label as part of the arc $(h,m,\ell)$.
This results in $|Labels| \times |Arcs|$ parts that need to be scored, leading
to slow parsing speeds and arguably a harder learning problem.

Instead, we chose to first predict the unlabeled structure using the model given
above, and then predict the label of each resulting arc.
Using this approach, the number of parts stays small, enabling fast parsing.

The labeling of an arc $(h,m)$ is performed using the same feature
representation $\phi(s,h,m)$ fed into a different MLP predictor:
\begin{align*}
    label(h,m) = \argmax_{\ell \in labels} MLP_{LBL}(v_h \circ v_m)[\ell]
\end{align*}
As before we use a margin based hinge loss.
The labeler is trained on the gold trees.\footnote{When training the labeled
parser, we calculate the structure loss and the labeling loss for each training
sentence, and sum the losses prior to computing the gradients.}
The \mbox{BiLSTM} encoder responsible for producing $v_h$ and $v_m$ is shared with the
arc-factored parser: the same \mbox{BiLSTM} encoder is used in the parer and the
labeler.  This
sharing of parameters can be seen as an instance of multi-task learning
\cite{Caruana:1997:ML:262868.262872}.
As we show in Section \ref{sec:results}, the sharing is effective: training the \mbox{BiLSTM} feature encoder to be good
at predicting arc-labels significantly improves the parser's unlabeled accuracy.

\paragraph{Loss augmented inference}
In initial experiments, the network learned quickly and overfit the data.
In order to remedy this, we found it useful to use \emph{loss augmented
inference} \cite{taskar2005learning}. %,crammer2006passive}.
The intuition behind loss augmented inference is to update against trees which
have high model scores and are also very wrong. This is done by augmenting the
score of each part not belonging to the gold tree by adding a constant to its score.
Formally, the loss transforms as follows:
\begin{align*}
&max(0, 1 + score(x,y) - \\
&\max_{y' \neq y} {\sum_{part \in y'} ( score_{local}(x,part) + \mathbbm{1}_{part \not \in y} ) } )
\end{align*}

\paragraph{Speed improvements}
The arc-factored model requires the scoring of $n^2$ arcs. Scoring is performed
using an MLP with one hidden layer, resulting in $n^2$ matrix-vector multiplications
from the input to the hidden layer, and $n^2$ multiplications from the hidden to
the output layer. The first $n^2$ multiplications involve larger dimensional
input and output vectors, and are the most time consuming. Fortunately, these
can be reduced to $2n$ multiplications and $n^2$ vector additions,
by observing that the multiplication
$W \cdot (v_h \circ v_m)$ can be written as $W^1 \cdot v_h + W^2 \cdot v_m$ where
$W^1$ and $W^1$ are are the first and second half of the matrix $W$ and reusing
the products across different pairs.

\noindent\textbf{Summary}
The graph-based parser is straight-forward first-order parser, trained with a
margin-based hinge-loss and loss-augmented inference. 
We depart from the literature by replacing the hand-crafted feature function
with a concatenation of \mbox{BiLSTM} representations of the head and modifier words,
and training the \mbox{BiLSTM} encoder jointly with the structured objective.
We also introduce a novel multi-task learning approach for labeled parsing %in which we
by training a second-stage arc-labeler sharing the same \mbox{BiLSTM} encoder with the unlabeled
parser.

\section{Experiments and Results}
\label{sec:results}

\begin{table*}[ht]
    \centering
\scalebox{0.74}{
\begin{tabular}{l|c|c|c|c|cc|cc}
    System & Method & Representation & Emb & PTB-YM & \multicolumn{2}{c}{PTB-SD} & \multicolumn{2}{c}{CTB} \\
           &        &                &     & UAS    & UAS & LAS & UAS & LAS \\
\hline
This work & graph, 1st order    & 2 \mbox{BiLSTM} vectors  & -- & -- & 93.1 & 91.0 & \textbf{86.6} & \textbf{85.1} \\
This work & transition (greedy, dyn-oracle) & 4 \mbox{BiLSTM} vectors  & -- & -- & 93.1 & 91.0 & 86.2 & 85.0 \\
This work & transition (greedy, dyn-oracle) & 11 \mbox{BiLSTM} vectors  & -- & -- & \textbf{93.2} & \textbf{91.2} & 86.5 & 84.9 \\
\hline
ZhangNivre11 & transition (beam) & large feature set (sparse) & -- & 92.9 & -- & -- & 86.0 & 84.4 \\
Martins13 (TurboParser) & graph, 3rd order+ & large feature set (sparse) & -- & 92.8 & 93.1 & -- & -- & -- \\
Pei15 & graph, 2nd order & large feature set (dense) & -- & 93.0 & -- & -- & -- & --\\
Dyer15 & transition (greedy) & Stack-LSTM + composition & -- & -- & 92.4 & 90.0 & 85.7 & 84.1 \\
Ballesteros16 & transition (greedy, dyn-oracle) & Stack-LSTM + composition & -- & -- & 92.7 & 90.6 & 86.1 & 84.5 \\
\hline\hline
This work & graph, 1st order    & 2 \mbox{BiLSTM} vectors  & YES & -- & 93.0 & 90.9 & 86.5 & 84.9 \\
This work & transition (greedy, dyn-oracle) & 4 \mbox{BiLSTM} vectors  & YES & -- & 93.6 & 91.5 & 87.4 & 85.9 \\
This work & transition (greedy, dyn-oracle) & 11 \mbox{BiLSTM} vectors  & YES & -- & 93.9 & 91.9 & \textbf{87.6} & 86.1 \\
\hline
Weiss15 & transition (greedy) & large feature set (dense) & YES & -- & 93.2 & 91.2 & -- & --\\
Weiss15 & transition (beam) & large feature set (dense) & YES & -- & \textbf{94.0} & \textbf{92.0} & -- & --\\
Pei15 & graph, 2nd order & large feature set (dense) & YES & 93.3 & -- & -- & -- & --\\
Dyer15 & transition (greedy) & Stack-LSTM + composition & YES & -- & 93.1 & 90.9 & 87.1 & 85.5 \\
Ballesteros16 & transition (greedy, dyn-oracle) & Stack-LSTM + composition & YES & -- & 93.6 & 91.4 & \textbf{87.6} & \textbf{86.2} \\
LeZuidema14 & reranking /blend & inside-outside recursive net & YES & 93.1 & 93.8 & 91.5 & -- & -- \\
Zhu15 & reranking /blend & recursive conv-net & YES & 93.8 & -- & -- & 85.7 & --\\
\end{tabular}}
\caption{\footnotesize Test-set parsing results of various state-of-the-art
parsing systems on the English (PTB) and Chinese (CTB) datasets. The systems
that use embeddings may use different pre-trained embeddings.
English results use predicted POS tags (different systems use different
taggers), while Chinese results use gold POS tags. 
\textbf{PTB-YM}: English PTB, Yamada and Matsumoto
head rules. \textbf{PTB-SD}: English PTB, Stanford Dependencies (different
systems may use different versions of the Stanford converter). \textbf{CTB}:
Chinese Treebank.  
\emph{reranking /blend} in Method column indicates a reranking system where the
reranker score is interpolated with the base-parser's score. 
The different systems and the numbers reported from them are taken from:
ZhangNivre11: \cite{zhang11acl}; Martins13: \cite{martins2013turbo}; Weiss15
\cite{weiss2015structured}; Pei15: \cite{pei2015effective}; Dyer15
\cite{dyer2015transitionbased}; Ballesteros16 \cite{ballesteros2016dynamic}; LeZuidema14
\cite{le2014insideoutside}; Zhu15: \cite{zhu2015reranking}. }
\label{tbl:sota}
\end{table*}

We evaluated our parsing model on English and Chinese data.
For comparison purposes we follow the setup of
\newcite{dyer2015transitionbased}.
\paragraph{Data}
For English, we used the Stanford Dependency (SD) \cite{sd} conversion of the Penn
Treebank \cite{penntb}, using the standard train/dev/test splits %\footnote{Training: secs 2-21. Development: sec 22. Test: sec 23.} 
with the same predicted POS-tags as used in
\newcite{dyer2015transitionbased};\newcite{chen2014fast}. This dataset contains a few non-projective trees.
Punctuation symbols are excluded from the evaluation.

For Chinese, we use the Penn Chinese Treebank 5.1 (CTB5), using the
train/test/dev splits of
\cite{tale-two-parsers,dyer2015transitionbased} %\footnote{Training: 001–815, 1001–1136. Development: 886–931, 1148–1151. Test: 816–885, 1137–1147.}
with gold
part-of-speech tags, also following
\cite{dyer2015transitionbased,chen2014fast}.

When using external word embeddings, we also use the same data as
\newcite{dyer2015transitionbased}.\footnote{We thank Dyer et al. for
sharing their data with us.}

\paragraph{Implementation Details}
The parsers are implemented in python, using the PyCNN
toolkit\footnote{\url{https://github.com/clab/cnn/tree/master/pycnn}} for neural network
training. The code is available at the github repository \url{https://github.com/elikip/bist-parser}.
We use the LSTM variant implemented in PyCNN, and optimize using the Adam
optimizer \cite{kingma2014adam}.  Unless otherwise noted, we use the default values provided
by PyCNN (e.g. for random initialization, learning rates etc).

The word and POS embeddings $e(w_i)$ and $e(p_i)$ are initialized to random
values and trained together with the rest of the parsers' networks. 
In some experiments, we introduce also pre-trained word embeddings. In those
cases,
the vector representation of a word is a concatenation
of its randomly-initialized vector embedding with its pre-trained word vector.
Both are tuned during training.
We use the same word vectors as in \newcite{dyer2015transitionbased}

During training, we employ a variant of \emph{word dropout}
\cite{iyyer2015deep}, and replace a word with the unknown-word symbol with probability
that is inversely proportional to the frequency of the word. A 
word $w$ appearing $\#(w)$ times in the training corpus is replaced with the unknown symbol
with probability  $p_{unk}(w) = \frac{\alpha}{\#(w)+\alpha}$.
If a word was dropped the external embedding of the word is also dropped with probability $0.5$.

We train the parsers for up to 30 iterations, and choose the best model
according to the UAS accuracy on the development set.

\paragraph{Hyperparameter Tuning}
We performed a very minimal hyper-parameter search with the graph-based parser,
and use the same hyper-parameters for both parsers.
The hyper-parameters of the final networks used for all the reported experiments
are detailed in Table \ref{tbl:hyper}.

\begin{table}[ht]
\begin{center}
\begin{scalebox}{0.8}{
\begin{tabular}{ | c | c | }
\hline 
Word embedding dimension & 100 \\
POS tag embedding dimension & 25 \\
Hidden units in $MLP$ & 100 \\
Hidden units in $MLP_{LBL}$ & 100 \\
BI-LSTM Layers & 2 \\
BI-LSTM Dimensions (hidden/output)  & 125 / 125 \\
$\alpha$ (for word dropout) & 0.25 \\
$p_{agg}$ (for exploration training) & 0.1 \\
\hline
\end{tabular}}
\end{scalebox}
\caption{Hyper-parameter values used in experiments}
\label{tbl:hyper}
\end{center}
\end{table}

\noindent\textbf{Main Results}
Table \ref{tbl:sota} lists the test-set accuracies of our best parsing models, compared
to other state-of-the-art parsers from the literature.\footnote{Unfortunately, many
papers still report English parsing results on the deficient Yamada and
Matsumoto head rules (PTB-YM) rather than the more modern Stanford-dependencies
(PTB-SD). We note that the PTB-YM and PTB-SD results are not strictly
comparable, and in our experience the PTB-YM results are usually about half a
UAS point higher.}

It is clear that our parsers are very competitive, despite using very simple
parsing architectures and minimal feature extractors.  
When not
using external embeddings, the first-order graph-based parser with 2 features outperforms
all other systems that are not using
external resources, including the third-order TurboParser. The greedy transition
based parser with 4 features also matches or outperforms most other parsers,
including the beam-based transition parser with heavily engineered features of Zhang and
Nivre (2011) and the Stack-LSTM parser of \newcite{dyer2015transitionbased}, as well as the same parser when trained using a dynamic oracle \cite{ballesteros2016dynamic}.
Moving from the simple (4 features) to the extended (11 features) feature set
leads to some gains in accuracy for both English and Chinese.

Interestingly, when adding external word embeddings the accuracy of the
graph-based parser \emph{degrades}.  We are not sure why this happens, and leave
the exploration of effective semi-supervised parsing with the graph-based model
for future work. The greedy parser does manage to benefit from the external
embeddings, and using them we also see gains from moving from the simple to
the extended feature set. Both feature sets result in very competitive results,
with the extended feature set yielding the best reported results for Chinese,
and ranked second for English, after the heavily-tuned beam-based parser
of \newcite{weiss2015structured}.

\paragraph{Additional Results}

We perform some ablation experiments in order to quantify the effect of the different
components on our best models (Table \ref{tbl:ablate}).

\begin{table}[h!]
\begin{center}
\begin{scalebox}{0.75}{
\begin{tabular}{l | c c | c c}
& \multicolumn{2}{c |}{PTB} & \multicolumn{2}{c}{CTB} \\
& UAS & LAS & UAS & LAS \\
\hline
Graph (no ext. emb) & 93.3 & 91.0 & 87.0 & 85.4 \\
 --POS & 92.9 & 89.8 & 80.6 & 76.8 \\
 --ArcLabeler & 92.7 & -- & 86.2 & -- \\
 --Loss Aug. & 81.3 & 79.4 & 52.6 & 51.7\\
%Graph:BiRNN & 89.0 & 85.9 & 82.4 & 80.5 \\
\hline
Greedy (ext. emb) & 93.8 & 91.5 & 87.8 & 86.0 \\
  --POS & 93.4 & 91.2 & 83.4 & 81.6 \\
  --DynOracle & 93.5 & 91.4 & 87.5 & 85.9 \\
%--Children & 93.0 & 91.0 & 86.5 & 84.9 \\
%--Children --Embedding & 92.8 & 90.5 & 86.0 & 84.2 \\
%--Oracle --Embedding & 93.0 & 90.7 & 85.7 & 84.1 \\
%Transition:BiRNN & 92.7 & 90.2 & 85.9 & 84.2 \\
\hline
\end{tabular}}
\end{scalebox}
\end{center}
\caption{\footnotesize Ablation experiments results (dev set) for the graph-based
parser without external embeddings and the greedy parser with external
embeddings and extended feature set.}
\label{tbl:ablate}
\end{table}

\vspace{5pt}
\noindent Loss augmented inference is crucial for the success of the graph-based
parser, and the multi-task learning scheme for the arc-labeler contributes nicely to
the \emph{unlabeled} scores. 
Dynamic oracle training 
yields nice gains for both English and Chinese.

\section{Conclusion}
We presented a pleasingly effective approach for feature extraction for dependency parsing
based on a \mbox{BiLSTM} encoder that is trained jointly with the parser, and
demonstrated its effectiveness by integrating it into two simple parsing models:
a greedy transition-based parser and a globally optimized first-order
graph-based parser, yielding very competitive parsing accuracies in both cases.

\paragraph{Acknowledgements}
This research is supported by the Intel Collaborative Research Institute for
Computational Intelligence (ICRI-CI) and the Israeli Science Foundation
(grant number 1555/15). 
We thank Lillian Lee for her important feedback and efforts invested in editing this paper. We also thank the reviewers for their valuable comments.

\bibliographystyle{acl}

\begin{thebibliography}{}

\bibitem[\protect\citename{Ballesteros \bgroup et al.\egroup
  }2016]{ballesteros2016dynamic}
Miguel Ballesteros, Yoav Goldberg, Chris Dyer, and Noah~A. Smith.
\newblock 2016.
\newblock Training with exploration improves a greedy stack-{LSTM} parser.
\newblock {\em CoRR}, abs/1603.03793.

\bibitem[\protect\citename{Caruana}1997]{Caruana:1997:ML:262868.262872}
Rich Caruana.
\newblock 1997.
\newblock Multitask learning.
\newblock {\em Machine Learning}, 28:41--75, July.

\bibitem[\protect\citename{Chen and Manning}2014]{chen2014fast}
Danqi Chen and Christopher Manning.
\newblock 2014.
\newblock A fast and accurate dependency parser using neural networks.
\newblock In {\em Proceedings of the 2014 Conference on Empirical Methods in
  Natural Language Processing (EMNLP)}, pages 740--750, Doha, Qatar, October.
  Association for Computational Linguistics.

\bibitem[\protect\citename{{Chiu} and {Nichols}}2016]{chiu2015named}
{Jason}~{P.C.} {Chiu} and {Eric} {Nichols}.
\newblock 2016.
\newblock Named entity recognition with bidirectional {LSTM-CNNs}.
\newblock {\em Transactions of the Association for Computational Linguistics},
  4.
\newblock To appear.

\bibitem[\protect\citename{Cho}2015]{cho-primer}
Kyunghyun Cho.
\newblock 2015.
\newblock Natural language understanding with distributed representation.
\newblock {\em CoRR}, abs/1511.07916.

\bibitem[\protect\citename{Cross and Huang}2016]{cross2016incremental}
James Cross and Liang Huang.
\newblock 2016.
\newblock Incremental parsing with minimal features using bi-directional
  {LSTM}.
\newblock In {\em Proceedings of the 54th Annual Meeting of the Association for
  Computational Linguistics}, Berlin, Germany, August. Association for
  Computational Linguistics.

\bibitem[\protect\citename{de Marneffe and Manning}2008]{sd}
Marie-Catherine de~Marneffe and Christopher~D. Manning.
\newblock 2008.
\newblock Stanford dependencies manual.
\newblock Technical report, {Stanford University}.

\bibitem[\protect\citename{Dyer \bgroup et al.\egroup
  }2015]{dyer2015transitionbased}
Chris Dyer, Miguel Ballesteros, Wang Ling, Austin Matthews, and Noah~A. Smith.
\newblock 2015.
\newblock Transition-based dependency parsing with stack long short-term
  memory.
\newblock In {\em Proceedings of the 53rd Annual Meeting of the Association for
  Computational Linguistics and the 7th International Joint Conference on
  Natural Language Processing (Volume 1: Long Papers)}, pages 334--343,
  Beijing, China, July. Association for Computational Linguistics.

\bibitem[\protect\citename{Eisner}1996]{eisner1996dep}
Jason Eisner.
\newblock 1996.
\newblock Three new probabilistic models for dependency parsing: An
  exploration.
\newblock In {\em 16th International Conference on Computational Linguistics,
  Proceedings of the Conference, {COLING} 1996, Center for Sprogteknologi,
  Copenhagen, Denmark, August 5-9, 1996}, pages 340--345.

\bibitem[\protect\citename{Elman}1990]{elman1990finding}
Jeffrey~L. Elman.
\newblock 1990.
\newblock Finding structure in time.
\newblock {\em Cognitive Science}, 14(2):179--211.

\bibitem[\protect\citename{Goldberg and Nivre}2012]{coling2012dynamic}
Yoav Goldberg and Joakim Nivre.
\newblock 2012.
\newblock A dynamic oracle for arc-eager dependency parsing.
\newblock In {\em Proceedings of COLING 2012}, pages 959--976, Mumbai, India,
  December. The COLING 2012 Organizing Committee.

\bibitem[\protect\citename{Goldberg and Nivre}2013]{tacl2013dynamic}
Yoav Goldberg and Joakim Nivre.
\newblock 2013.
\newblock Training deterministic parsers with non-deterministic oracles.
\newblock {\em Transactions of the Association for Computational Linguistics},
  1:403--414.

\bibitem[\protect\citename{Goldberg}2015]{goldberg-primer}
Yoav Goldberg.
\newblock 2015.
\newblock A primer on neural network models for natural language processing.
\newblock {\em CoRR}, abs/1510.00726.

\bibitem[\protect\citename{Graves}2008]{graves2008supervised}
Alex Graves.
\newblock 2008.
\newblock {\em Supervised sequence labelling with recurrent neural networks}.
\newblock {Ph.D.} thesis, Technical University Munich.

\bibitem[\protect\citename{Hochreiter and Schmidhuber}1997]{hochreiter1997long}
Sepp Hochreiter and J{\"{u}}rgen Schmidhuber.
\newblock 1997.
\newblock Long short-term memory.
\newblock {\em Neural Computation}, 9(8):1735--1780.

\bibitem[\protect\citename{Huang and Sagae}2010]{huang-sagae:2010:ACL}
Liang Huang and Kenji Sagae.
\newblock 2010.
\newblock Dynamic programming for linear-time incremental parsing.
\newblock In {\em Proceedings of the 48th Annual Meeting of the Association for
  Computational Linguistics}, pages 1077--1086, Uppsala, Sweden, July.
  Association for Computational Linguistics.

\bibitem[\protect\citename{Irsoy and Cardie}2014]{irsoy2014opinion}
Ozan Irsoy and Claire Cardie.
\newblock 2014.
\newblock Opinion mining with deep recurrent neural networks.
\newblock In {\em Proceedings of the 2014 Conference on Empirical Methods in
  Natural Language Processing (EMNLP)}, pages 720--728, Doha, Qatar, October.
  Association for Computational Linguistics.

\bibitem[\protect\citename{Iyyer \bgroup et al.\egroup }2015]{iyyer2015deep}
Mohit Iyyer, Varun Manjunatha, Jordan Boyd-Graber, and Hal Daum\'{e}~III.
\newblock 2015.
\newblock Deep unordered composition rivals syntactic methods for text
  classification.
\newblock In {\em Proceedings of the 53rd Annual Meeting of the Association for
  Computational Linguistics and the 7th International Joint Conference on
  Natural Language Processing (Volume 1: Long Papers)}, pages 1681--1691,
  Beijing, China, July. Association for Computational Linguistics.

\bibitem[\protect\citename{Karpathy \bgroup et al.\egroup
  }2015]{karpathy2015visualizing}
Andrej Karpathy, Justin Johnson, and Fei{-}Fei Li.
\newblock 2015.
\newblock Visualizing and understanding recurrent networks.
\newblock {\em CoRR}, abs/1506.02078.

\bibitem[\protect\citename{Kingma and Ba}2015]{kingma2014adam}
Diederik~P. Kingma and Jimmy Ba.
\newblock 2015.
\newblock Adam: {A} method for stochastic optimization.
\newblock In {\em Proceedings of the 3rd International Conference for Learning
  Representations}, San Diego, California.

\bibitem[\protect\citename{Kiperwasser and Goldberg}2016]{kiperwasser2016ef}
Eliyahu Kiperwasser and Yoav Goldberg.
\newblock 2016.
\newblock Easy-first dependency parsing with hierarchical tree {LSTMs}.
\newblock {\em Transactions of the Association for Computational Linguistics},
  4.
\newblock To appear.

\bibitem[\protect\citename{Koo and Collins}2010]{koo2010thirdorder}
Terry Koo and Michael Collins.
\newblock 2010.
\newblock Efficient third-order dependency parsers.
\newblock In {\em Proceedings of the 48th Annual Meeting of the Association for
  Computational Linguistics}, pages 1--11, Uppsala, Sweden, July. Association
  for Computational Linguistics.

\bibitem[\protect\citename{Koo \bgroup et al.\egroup }2008]{koo2008semisup}
Terry Koo, Xavier Carreras, and Michael Collins.
\newblock 2008.
\newblock Simple semi-supervised dependency parsing.
\newblock In {\em Proceedings of the 46th Annual Meeting of the Association for
  Computational Linguistics}, pages 595--603, Columbus, Ohio, June. Association
  for Computational Linguistics.

\bibitem[\protect\citename{K{\"{u}}bler \bgroup et al.\egroup
  }2009]{kubler2008dependency}
Sandra K{\"{u}}bler, Ryan~T. McDonald, and Joakim Nivre.
\newblock 2009.
\newblock {\em Dependency Parsing}.
\newblock Synthesis Lectures on Human Language Technologies. Morgan {\&}
  Claypool Publishers.

\bibitem[\protect\citename{Kuhlmann \bgroup et al.\egroup }2011]{kuhlmann2011}
Marco Kuhlmann, Carlos G\'{o}mez-Rodr\'{i}guez, and Giorgio Satta.
\newblock 2011.
\newblock Dynamic programming algorithms for transition-based dependency
  parsers.
\newblock In {\em Proceedings of the 49th Annual Meeting of the Association for
  Computational Linguistics: Human Language Technologies}, pages 673--682,
  Portland, Oregon, USA, June. Association for Computational Linguistics.

\bibitem[\protect\citename{Lample \bgroup et al.\egroup
  }2016]{lample2016neural}
Guillaume Lample, Miguel Ballesteros, Sandeep Subramanian, Kazuya Kawakami, and
  Chris Dyer.
\newblock 2016.
\newblock Neural architectures for named entity recognition.
\newblock In {\em Proceedings of the 2016 Conference of the North American
  Chapter of the Association for Computational Linguistics: Human Language
  Technologies}, pages 260--270, San Diego, California, June. Association for
  Computational Linguistics.

\bibitem[\protect\citename{Le and Zuidema}2014]{le2014insideoutside}
Phong Le and Willem Zuidema.
\newblock 2014.
\newblock The inside-outside recursive neural network model for dependency
  parsing.
\newblock In {\em Proceedings of the 2014 Conference on Empirical Methods in
  Natural Language Processing (EMNLP)}, pages 729--739, Doha, Qatar, October.
  Association for Computational Linguistics.

\bibitem[\protect\citename{LeCun \bgroup et al.\egroup
  }2006]{lecun2006tutorial}
Yann LeCun, Sumit Chopra, Raia Hadsell, Marc’Aurelio Ranzato, and Fu~Jie
  Huang.
\newblock 2006.
\newblock A tutorial on energy-based learning.
\newblock {\em Predicting structured data}, 1.

\bibitem[\protect\citename{Lei \bgroup et al.\egroup
  }2014]{lei-EtAl:2014:P14-1}
Tao Lei, Yu~Xin, Yuan Zhang, Regina Barzilay, and Tommi Jaakkola.
\newblock 2014.
\newblock Low-rank tensors for scoring dependency structures.
\newblock In {\em Proceedings of the 52nd Annual Meeting of the Association for
  Computational Linguistics (Volume 1: Long Papers)}, pages 1381--1391,
  Baltimore, Maryland, June. Association for Computational Linguistics.

\bibitem[\protect\citename{Lewis \bgroup et al.\egroup }2016]{lewis2016lstm}
Mike Lewis, Kenton Lee, and Luke Zettlemoyer.
\newblock 2016.
\newblock {LSTM} {CCG} parsing.
\newblock In {\em Proceedings of the 2016 Conference of the North American
  Chapter of the Association for Computational Linguistics: Human Language
  Technologies}, pages 221--231, San Diego, California, June. Association for
  Computational Linguistics.

\bibitem[\protect\citename{Marcus \bgroup et al.\egroup }1993]{penntb}
Mitchell~P. Marcus, Beatrice Santorini, and Mary~Ann Marcinkiewicz.
\newblock 1993.
\newblock Building a large annotated corpus of {E}nglish: The {P}enn
  {T}reebank.
\newblock {\em Computational Linguistics}, 19(2):313--330.

\bibitem[\protect\citename{Martins \bgroup et al.\egroup }2009]{martins2009ilp}
Andre Martins, Noah~A. Smith, and Eric Xing.
\newblock 2009.
\newblock Concise integer linear programming formulations for dependency
  parsing.
\newblock In {\em Proceedings of the Joint Conference of the 47th Annual
  Meeting of the ACL and the 4th International Joint Conference on Natural
  Language Processing of the AFNLP}, pages 342--350, Suntec, Singapore, August.
  Association for Computational Linguistics.

\bibitem[\protect\citename{Martins \bgroup et al.\egroup
  }2013]{martins2013turbo}
Andre Martins, Miguel Almeida, and Noah~A. Smith.
\newblock 2013.
\newblock Turning on the turbo: Fast third-order non-projective turbo parsers.
\newblock In {\em Proceedings of the 51st Annual Meeting of the Association for
  Computational Linguistics (Volume 2: Short Papers)}, pages 617--622, Sofia,
  Bulgaria, August. Association for Computational Linguistics.

\bibitem[\protect\citename{McDonald \bgroup et al.\egroup }2005]{mst}
Ryan McDonald, Koby Crammer, and Fernando Pereira.
\newblock 2005.
\newblock Online large-margin training of dependency parsers.
\newblock In {\em Proceedings of the 43rd Annual Meeting of the Association for
  Computational Linguistics (ACL'05)}, pages 91--98, Ann Arbor, Michigan, June.
  Association for Computational Linguistics.

\bibitem[\protect\citename{McDonald}2006]{mcdonald-phd}
Ryan McDonald.
\newblock 2006.
\newblock {\em Discriminative Training and Spanning Tree Algorithms for
  Dependency Parsing}.
\newblock {Ph.D.} thesis, University of Pennsylvania.

\bibitem[\protect\citename{Nivre}2004]{arc-eager}
Joakim Nivre.
\newblock 2004.
\newblock Incrementality in deterministic dependency parsing.
\newblock In Frank Keller, Stephen Clark, Matthew Crocker, and Mark Steedman,
  editors, {\em Proceedings of the ACL Workshop Incremental Parsing: Bringing
  Engineering and Cognition Together}, pages 50--57, Barcelona, Spain, July.
  Association for Computational Linguistics.

\bibitem[\protect\citename{Nivre}2008]{nivre2008algorithms}
Joakim Nivre.
\newblock 2008.
\newblock Algorithms for deterministic incremental dependency parsing.
\newblock {\em Computational Linguistics}, 34(4):513--553.

\bibitem[\protect\citename{Pei \bgroup et al.\egroup }2015]{pei2015effective}
Wenzhe Pei, Tao Ge, and Baobao Chang.
\newblock 2015.
\newblock An effective neural network model for graph-based dependency parsing.
\newblock In {\em Proceedings of the 53rd Annual Meeting of the Association for
  Computational Linguistics and the 7th International Joint Conference on
  Natural Language Processing (Volume 1: Long Papers)}, pages 313--322,
  Beijing, China, July. Association for Computational Linguistics.

\bibitem[\protect\citename{Schuster and
  Paliwal}1997]{schuster1997bidirectional}
Mike Schuster and Kuldip~K. Paliwal.
\newblock 1997.
\newblock Bidirectional recurrent neural networks.
\newblock {\em {IEEE} Trans. Signal Processing}, 45(11):2673--2681.

\bibitem[\protect\citename{Taskar \bgroup et al.\egroup
  }2005]{taskar2005learning}
Benjamin Taskar, Vassil Chatalbashev, Daphne Koller, and Carlos Guestrin.
\newblock 2005.
\newblock Learning structured prediction models: A large margin approach.
\newblock In {\em Machine Learning, Proceedings of the Twenty-Second
  International Conference {(ICML} 2005), Bonn, Germany, August 7-11, 2005},
  pages 896--903.

\bibitem[\protect\citename{Taub-Tabib \bgroup et al.\egroup
  }2015]{taubtabib2015template}
Hillel Taub-Tabib, Yoav Goldberg, and Amir Globerson.
\newblock 2015.
\newblock Template kernels for dependency parsing.
\newblock In {\em Proceedings of the 2015 Conference of the North American
  Chapter of the Association for Computational Linguistics: Human Language
  Technologies}, pages 1422--1427, Denver, Colorado, May--June. Association for
  Computational Linguistics.

\bibitem[\protect\citename{Titov and
  Henderson}2007]{titov-henderson:2007:IWPT2007}
Ivan Titov and James Henderson.
\newblock 2007.
\newblock A latent variable model for generative dependency parsing.
\newblock In {\em Proceedings of the Tenth International Conference on Parsing
  Technologies}, pages 144--155, Prague, Czech Republic, June. Association for
  Computational Linguistics.

\bibitem[\protect\citename{Vaswani \bgroup et al.\egroup
  }2016]{Vaswani:2016:NAACL}
Ashish Vaswani, Yonatan Bisk, Kenji Sagae, and Ryan Musa.
\newblock 2016.
\newblock Supertagging with {LSTMs}.
\newblock In {\em Proceedings of the 15th Annual Conference of the North
  American Chapter of the Association for Computational Linguistics (Short
  Papers)}, San Diego, California, June.

\bibitem[\protect\citename{Vinyals \bgroup et al.\egroup
  }2015]{vinlays2014grammar}
Oriol Vinyals, Lukasz Kaiser, Terry Koo, Slav Petrov, Ilya Sutskever, and
  Geoffrey~E. Hinton.
\newblock 2015.
\newblock Grammar as a foreign language.
\newblock In {\em Advances in Neural Information Processing Systems 28: Annual
  Conference on Neural Information Processing Systems 2015, December 7-12,
  2015, Montreal, Quebec, Canada}, pages 2773--2781.

\bibitem[\protect\citename{Weiss \bgroup et al.\egroup
  }2015]{weiss2015structured}
David Weiss, Chris Alberti, Michael Collins, and Slav Petrov.
\newblock 2015.
\newblock Structured training for neural network transition-based parsing.
\newblock In {\em Proceedings of the 53rd Annual Meeting of the Association for
  Computational Linguistics and the 7th International Joint Conference on
  Natural Language Processing (Volume 1: Long Papers)}, pages 323--333,
  Beijing, China, July. Association for Computational Linguistics.

\bibitem[\protect\citename{Zhang and Clark}2008]{tale-two-parsers}
Yue Zhang and Stephen Clark.
\newblock 2008.
\newblock A tale of two parsers: {I}nvestigating and combining graph-based and
  transition-based dependency parsing.
\newblock In {\em Proceedings of the 2008 Conference on Empirical Methods in
  Natural Language Processing}, pages 562--571, Honolulu, Hawaii, October.
  Association for Computational Linguistics.

\bibitem[\protect\citename{Zhang and Nivre}2011]{zhang11acl}
Yue Zhang and Joakim Nivre.
\newblock 2011.
\newblock Transition-based dependency parsing with rich non-local features.
\newblock In {\em Proceedings of the 49th Annual Meeting of the Association for
  Computational Linguistics: Human Language Technologies}, pages 188--193,
  Portland, Oregon, USA, June. Association for Computational Linguistics.

\bibitem[\protect\citename{Zhu \bgroup et al.\egroup }2015]{zhu2015reranking}
Chenxi Zhu, Xipeng Qiu, Xinchi Chen, and Xuanjing Huang.
\newblock 2015.
\newblock A re-ranking model for dependency parser with recursive convolutional
  neural network.
\newblock In {\em Proceedings of the 53rd Annual Meeting of the Association for
  Computational Linguistics and the 7th International Joint Conference on
  Natural Language Processing (Volume 1: Long Papers)}, pages 1159--1168,
  Beijing, China, July. Association for Computational Linguistics.

\end{thebibliography}

\clearpage

\end{document}